\documentclass[prl,amsmath,twocolumn,showpacs]{revtex4}
\usepackage{graphicx}
\usepackage{amsfonts}
\usepackage{amsmath}
\usepackage{amssymb}
\usepackage{epsfig}
\usepackage{epstopdf}
\usepackage{upgreek}
\usepackage[usenames,dvipsnames]{color}
\begin{document}
\title{L\'{e}vy robotics}

\author{M. Krivonosov}
\affiliation{Department of Applied Mathematics, Lobachevsky State University of Nizhny Novgorod, 603140 N. Novgorod, Russia}
\author{S. Denisov}
\affiliation{Department of Applied Mathematics, Lobachevsky State University of Nizhny Novgorod, 603140 N. Novgorod, Russia}
\affiliation{Institute of Physics, University of Augsburg, Universitätsstrasse 1, D-86135, Augsburg Germany}
\author{V. Zaburdaev}
\affiliation{Institute of Supercomputing Technologies, Lobachevsky State University of Nizhny Novgorod, 603140 N. Novgorod, Russia}
\affiliation{Max Planck Institute for the Physics of Complex Systems, N\"{o}thnitzer Str. 38, D-01187 Dresden, Germany}

\begin{abstract}
Two the most common tasks for autonomous mobile robots is to explore the environment and locate a target. 
Targets could range from sources of chemical contamination to people 
needing assistance in a disaster area. From the very beginning, the quest for most efficient search algorithms was strongly influenced by behavioral science and ecology,
where researchers try to figure out the strategies used by leaving beings, from bacteria to mammals. Since then, bio-inspired random search algorithms remain one the most important 
directions in autonomous robotics. Recently a new wave arrived bringing a specific type of random walks as a universal  search
strategy exploited by immune cells, insects,  mussels, albatrosses, sharks, deers, and a dozen of other animals including humans. 
These \textit{L\'{e}vy} walks combine two key features, the ability of walkers to spread  anomalously fast while moving with a finite velocity.
The latter is especially valuable in the context of robotics  because it respects  the reality autonomous robots live in. There is already an
impressive body of publications on L\'{e}vy robotics; yet research in this field is unfolding further, constantly 
bringing  new results, ideas, hypothesis and speculations. In this mini-review  we survey the current state of the field, 
list latest advances, discuss the prevailing trends, and outline further perspectives.
\end{abstract}

\maketitle

\section{Introduction}\label{Introduction}

The problem of target search by mobile autonomous robots \cite{0} is a  
research topic which remains hot for already several decades. 
The performance of the robot (or, equivalently, the efficiency of the corresponding algorithm) can be 
quantified, depending on the context, by using either the time it takes the robot to find a specific target
or by the number of targets robot finds for a fixed time ('foraging' \cite{2}). There are two ultimate alternatives of search, that are
random and deterministic searches \cite{1}. No  information is available about the location
of targets in the case of a pure random search -- until the robot simply hits it. When performing a pure  deterministic search, the robot 
can 'sense' the target from a distance, by mounted sensors, and just moves towards it. However, there is a continuous spectrum of strategies
in between these two extremes \cite{3}. In planar environments where the distribution of targets
is unknown a priori or changes over time randomized search strategies were suggested to be more efficient \cite{4}.

We can assume that the robot moves with a constant velocity. This simple,  intuitive, and, in fact, legit assumption allows us to handle, 
in one go,  such important issues as energy cost of the movement and continuity of the  motion. 
Then an optimal search strategies can be defined as that minimizing  the mean time needed for a robot to hit a  target.
This condition is identical  to minimization of  the expenditure of energy along the way. In the case of random search, optimization 
could only be achieved by tuning the parameters of robot's motion. 

Animals perform random search almost every day of their lives. Often they do it
with no knowledge of the environment and a limited sensory range. However, they were given thousands of years to optimize their search (a proper name in this context is 'foraging) strategies \cite{5}.
They also pay an energy price for their motion, their movements are continuous and usually performed with some more or less constant velocity (at least, during the search phase). 
Shortly, they are quite similar to robots in this respect. In this situation the idea of “learning from nature” is beneficial. 

\begin{figure*}[t]
\begin{tabular}{cc}
\includegraphics[width=0.9\textwidth]{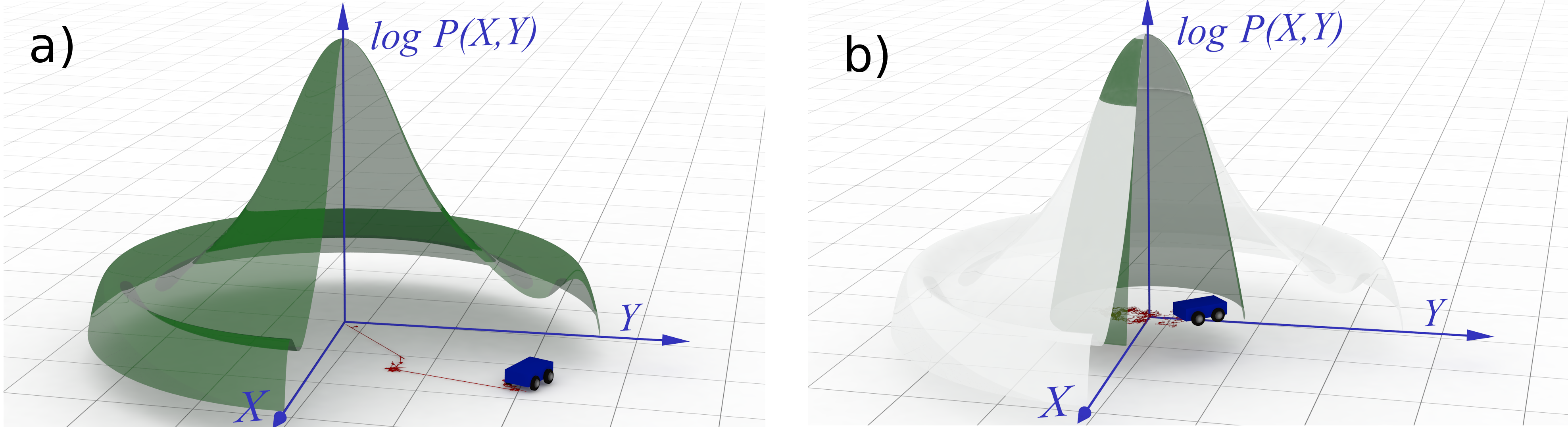}
\end{tabular}
\caption{(a) Robots performing planar L\'{e}vy walks (a) with exponent $\gamma = 1.2$, Eq.~(1), and a standard Pearson random walk (b), $\psi(\tau) \equiv 1$.
Both robots move with a constant (by absolute value) velocity $\upsilon$.
The difference between the two motion strategies can be seen from the probability density function $P(x,y;t)$ to find
a robot at specific point $\{X,Y\}$ at time $t$ under the condition that the robot was launched from the point $\{0,0\}$.
The pdfs look similar  near the origin; it is always possible to fit the hat of the Levy robot's pdf with a Gaussian distribution
of a some finite variance. However, away of the origin L\'{e}vy pdf  has a distinctive slow-decaying power-law tail. 
Note that probability to reach a point located on the ballistic front, $| \mathbf{r}_{\mathrm{bal}}(t)| = \sqrt{X(t)^2 +Y(t)^2} = \upsilon t$, is non-zero in both cases,
yet in the case of Pearson walk it is negligible on long-time scale simply because the  Gaussian function decays super-exponentially.}
\label{fig1}
\end{figure*}

During the last  two decades, it has been noticed that the foraging trajectories  of many
animals, ranging from honeybees \cite{honey} to sharks \cite{shark} and  human gatherers \cite{gazda}, appear to be very similar.  Namely, 
they all look like  periods of localized diffusive-like search activity altered with ballistic relocations to a new spot \cite{benya}. 
Such intermittent trajectories
are a trademark of the so-called 'L\'{e}vy walks' (LW) \cite{LW}, which, in fact, have no characteristic length scale \cite{yossi}.  Indeed, the statistical analysis
of the experimental trajectories uncovered the main common feature: the length  distribution of 'ballistic relocations' is well  approximated by a power-law distribution
$\varphi (l) \propto l^{-1-\gamma}$, with the exponent $1 <  \gamma < 2$. There is only a one-step distance from this observation to the concept of L\'{e}vy walks, 
namely a constant (by absolute value) velocity of  motion, $\upsilon$, during each relocation step \cite{LW0} (generalizations of the original model with non-constant but always finite velocities are collected  in Ref.~\cite{LW}).
Viswanathan et al. \cite{wis} has considered a L\'{e}vy walker which is wandering over an infinite  plane and constantly 
searching for targets, randomly distributed in space with a unifrom density.
The set-up falls in the category of what described above as random search. Viswanathan et al. considered two possible scenarios, 
with  targets destroyed after being found (destructive case) or becaming temporally depleted  (non-destructive case). The main finding was that 
the L\'{e}vy walk with exponent $\gamma =2$ turned to be most efficient, as compared to other LWs and simple ballistic motion. This result
greatly promoted  LWs as an optimal search strategy and generated  a whole wave of research activity. 
We will not survey here the still growing  body of publications (at the moment about a hundred) on L\'{e}vy foraging and search; we refer the interested reader to the available reviews and monographs. 
It is noteworthy, however,  that the overall attitude in these fields has changed from the enthusiasms to a bona fide  skepticism and 
a sober critical view on the LW-oriented  interpretations of the collected statistical data; see a recent work~\cite{crit2} and  a series of commenting publications initiated by it.

It is not a surprise then that the wave of studies on L\'{e}vy foraging and animal search strategies has attracted attention of the researchers working in the field of robotics. We have noticed 
two  trends in L\'{e}vy robotics, which are complementary to one another.  First one goes along the line ``learn from the Nature'' we discussed above
and deals with the development of new LW-inspired search algorithms for autonomous mobile robots \cite{Pasternak2009,Nurzaman2009,Lenagh2010,Sutantyo2010,Fujisawa2013,Sutantyo2013}. 
Second, a relatively recent one, aims  at the
understanding of how L\'{e}vy-walk motion patterns emerge from combinations of different external factors and also used as a test to probe 
theoretical assumptions on animal strategies  \cite{Fricke2013}.

In this small review we survey the filed of  L\'{e}vy robotics as it stands by now. To make it self-sufficient, we start with Section 1 introducing
the LW concept. In Section 2 we discuss some important (in our opinion) publications on the subject. In the concluding section we outline some perspectives
and discuss potentially interesting problems for us, LW theoreticians, motivated by what we have learned from the publications.

\section{L\'{e}vy walk summary}
The definition of a simplest L\'{e}vy walk model on a two-dimensional plane is very close to the original formulation 
of the random walk given more than a hundred years ago by Pearson \cite{pearson}. A walker chooses a random direction and a 
random time $\tau$ and walks with a constant speed $\upsilon$ in the selected direction. After the time has elapsed a new random 
direction and a new random time are picked and the process repeats. Importantly, the durations of walks are distributed according to a power-law density:
\begin{equation}
\psi(\tau)=\frac{1}{\tau_0}\frac{\gamma}{(1+\tau/\tau_0)^{1+\gamma}}, \gamma>0
\end{equation}
Particular details of this distribution are not as important, but its slowly decaying power-law tail is 
very central in determining the dispersal process on long time scales. Different values of $\gamma$ correspond to different 
regimes of the dispersal. Such for $\gamma>2$ the mean squared step length [calculated simply as $\langle v^2\tau^2\rangle$] is 
finite and the Central Limit Theorem CLT predicts normal Gaussian diffusion as an outcome of such a walk. Situation changes 
when $\gamma$ drops below 2. The mean squared step size becomes infinite and the CLT breaks down. Instead, there is a generalized 
CLT saying that in this case the spatial distribution of walkers should look like a L\'{e}vy distribution (hence the name of the walk). 
The hallmark of the L\'{e}vy distribution are slowly decaying power-law tails which clearly set it apart from the Gaussian profile 
with super-exponentially fast vanishing tails, see Fig.~1. The tails, however, do not spread to infinity, but are cut off by the ballistic front -- 
at any moment of time $t$ there can be no particles beyond the line $|\mathbf{r}_{\mathrm{bal}}| =  \upsilon t$ (which often is referred to as a 'light cone'). 
As more quantitative information we mention that the density of particles in the bulk of the distribution follows the scaling $x\propto t^{1/\gamma}$ 
(one can understand that as the characteristic scale of the particle cloud) and the mean squared displacement, a standard measure of how far are 
particles from the starting point, behaves as $\langle\mathbf{r^2}\rangle\propto t^{3-\gamma}$. For $\gamma<1$ even more dramatic things happen 
in that the density of particles looks more like a well instead of a hump and exhibits ballistic scaling for the whole distribution. As currently there are not so 
many known real world examples of the ballistic regime, we will not consider it further, but they are worthy of being remembered as a principal possibility in the context of robotics.

\section{L\'{e}vy walks in autonomous robotics}\label{main}

A first idea to combine L\'{e}vy walks with chemotaxis in order to obtain  a search 
algorithm for an autonomous agent to find a source of chemical contamination in a turbulent aquatic environment, was proposed by Pasternak \textit{et al.}
\cite{Pasternak2009}. It is not a
typical search task because the searcher should scan a constantly changing 
chemical field and follow plumes in order to find their origin. In the computational studies, 
a virtual AUV (Autonomous Underwater Vehicle), floating in a virtual two-dimensional river-like
turbulent flow, contaminated from a  point-like source, was used. Events of unidirectional motion,
characterized by a power-law distribution of their lengths and a wrapped Cauchy distribution of 
their direction angles, were intermingled with short re-orientation events. During the latter
the vehicle was randomly choosing a new movement direction 
along the local concentration upstream flow. 
This strategy somehow corresponds to a L\'{e}vy walk in a flow-oriented reference frame. 
When compared to other strategies, based on 
Brownian walk, simple L\'{e}vy walk, correlated Brownian walk and a brute-force zig-zag scanning, 
L\'{e}vy-taxis outperformed all of them, both in terms of detection success rate and detection speed.  

Another searching strategy for a mobile robot, a sequence of L\'{e}vy walks alternated with taxis 
events, was proposed by Nurzaman \textit{et al.} \cite{Nurzaman2009}. In computer simulations, the robot task was to locate a 
loudspeaker by using the information 
on the local sound intensity obtained from a robot-mounted microphone. The loudspeaker was stationary and the robot's 
speed $\upsilon$ was constant.
\begin{figure}[t]
\center
\includegraphics[width=0.45\textwidth]{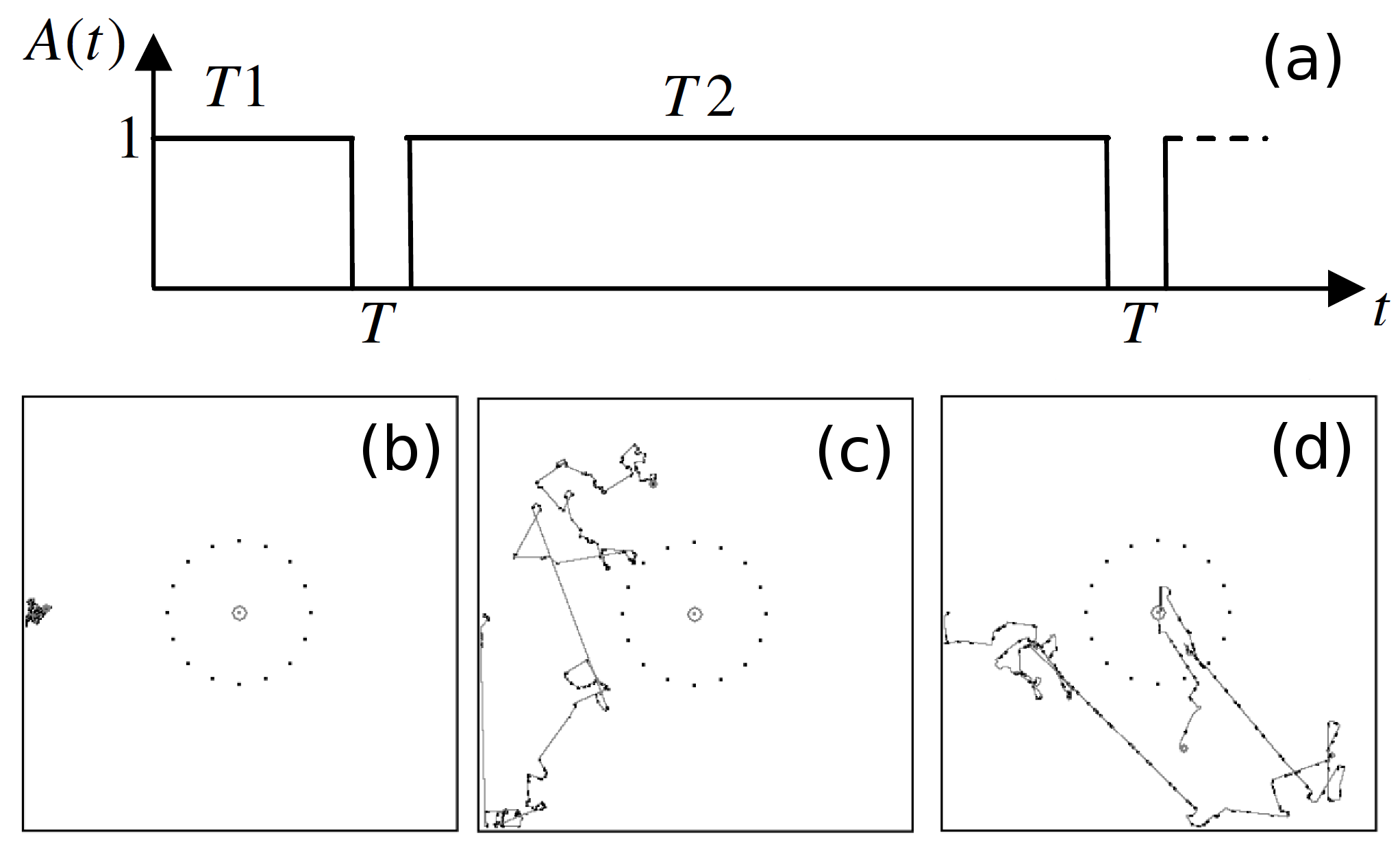}
\caption[robot1]
{Performance of a sonotactic robot. (a) Activity of the robot: Durations $T1,~T2,...$ follow a power-law tail
distribution while the duration of  re-orientation events $T$ is constant; (b-d) Trajectories of the robot using the sonotaxis strategy (b),
the L\'{e}vy walk (c) and the combination of the two (d). Speaker (small solid circle) is located at the center of the squared 
test area (box) and dashed line encircles the area with sound gradient above a threshold. The starting point is located at 
the middle of the left box border.  Figure courtesy of S. G. Nurzaman.}
\label{robot1}
\end{figure}
The robot orientation was defined by the angle $\theta$. The robot dynamics 
was governed by three stochastic equations,
\begin{eqnarray}
\left[\!\!\begin{array}{cc}
                  \dot{x}(t)\\
                \dot{y}(t)\\
                  \dot{\theta}(t)
             \end{array}
              \!\!\right] =
A(t)\left[\!\!\begin{array}{cc}
                  \upsilon\cos\theta(t)\\
                \upsilon\sin\theta(t)\\
                 0
             \end{array}
              \!\!\right]+
[1-A(t)]\left[\!\!\begin{array}{cc}
                  0\\
                0\\
                 \varepsilon_{\theta}(t)
             \end{array}
              \!\!\right]
\label{robot_formula}
\end{eqnarray}
where the Cartesian coordinates $x(t)$ and $y(t)$ specify the position of the robot at time $t$. 
Activity $A(t)$  is  a dichotomous function switching between $1$ and $0$ so that the robot
is either moving forward with velocity $\upsilon$ (activity is ``1'') or is choosing randomly a new direction of motion
(activity is ``0''). When the duration of a single  $1$-event is distributed according to a power-law, 
see Fig.~\ref{robot1}(a), the robot performs a two-dimensional version of the L\'{e}vy walk with rests shown on Fig.~\ref{fig1}(a).  
Alternatively,  a stochastic sonotaxis strategy by using which the robot tried to locate  and move towards the loudspeaker was probed. 
However, neither of the two strategies was able to accomplish the task when used alone.
The sonotaxis turned out to be effective in a close vicinity of the speaker only, 
and did not work when the sound gradient was small, see Fig.~\ref{robot1}(a). The L\'{e}vy walk did not care about the sound 
intensity by default and produced unbiased wandering only, Fig.~\ref{robot1}(b). 
The combination of the two solved the problem: 
the L\'{e}vy walk first brought the robot to the area where the sound-intensity gradient  was high enough and from there
the sonotaxis strategy was able to lead the robot to the loudspeaker, Fig.~\ref{robot1}(c).
A \textit{L\'{e}vy looped search} algorithm  to locate \textit{mobile} targets with a swarm of non-interacting  robots
was proposed by Lenagh and Dasgupta \cite{Lenagh2010}. The idea was to replace straight ballistic segments with loops so that each searcher returns to its 
initial position. The length of each loop was sampled from a power-law distribution, whereas the starting angle 
was sampled from the uniform distribution in the interval $[0, 2\pi]$. The reported results showed that the looped search 
outperformed the standard L\'{e}vy search in tracking mobile targets.
\begin{figure}[t]
\center
\includegraphics[width=0.5\textwidth]{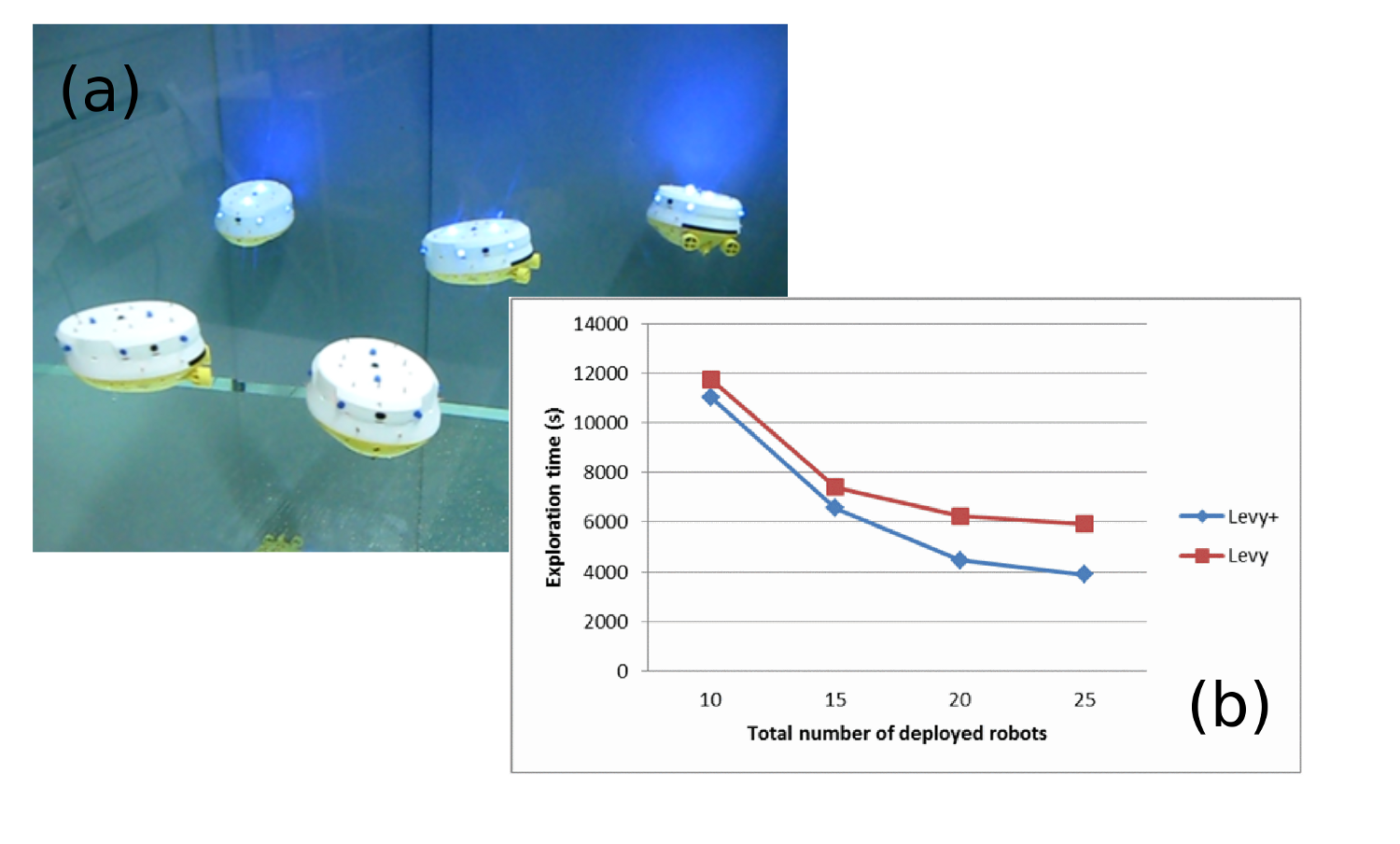}
\caption[robot2]
{Collective  multi-robot exploration. (a) Autonomous underwater vehicles used in the experiments; (b)
Targets searching experimental results: Exploration time vs number of robots for two strategies, 
with independent L\'{e}vy searchers (red line) and and interacting L\'{e}vy searchers (blue line). 
Adapted from Ref.~\cite{Sutantyo2013}.}
\label{robot2}
\end{figure}

The idea that a search efficiency can be increased by using a number of autonomous agents is natural and
relevant in many contexts. It is evident, for example, that the search time is inversely proportional to the number 
of independent searchers provided all  
other conditions remain the same. However, if an interaction or exchange of information between the searchers is allowed,
the search time can be decreased even further. Swarm communication is widely used among animals and insects, 
and it is known  among biologists and roboticists as ``stigmergy'' \cite{stigmergy}. A multi-robot searching algorithm based 
on a combination of a L\'{e}vy walk and an artificial potential field inducing repulsion among robots, was proposed and tested 
by Sutantyo \textit{et al.} \cite{Sutantyo2010}. The obtained results for up to twenty robots showed that the repulsion increases search efficiency 
in terms of the search time. It is noteworthy that the effect diminishes with increase of the robot number,  because
crowding robots start to change their directions  earlier than expected from the governing power-law distribution. 
Experimental results obtained for 
two L\'{e}vy-swimming AUVs in a $3$-d  aquatic testbed \cite{Keteer2012} show that in this case the best performance corresponds 
to a simple divide-and-conquer strategy, when the tank is divided into two equal volumes and each submarine scouts its assigned region only.
However this situation may change when the number of AUVs is larger than two so that communication between searchers could be beneficial.
Group L\'{e}vy foraging with an artificial pheromone communication between  robots was studied recently by Fujisawa and Dobata \cite{Fujisawa2013}. 
Each robot had a tank filled with a ``pheromone'' (alcohol)  which was sprayed around by a micropump. 
Rovers also carried alcohol and touch sensors and  
their motion was controlled  by a program which 
took into account the local pheromone concentration. 
The swarm foraging efficiency  peaked when the robots were  programmed beforehand  to perform a L\'{e}vy walk  in the
absence of the communication. Multi-robot underwater exploration and target location were studied
with a swarm of L\'{e}vy-swimming AUVs by Sutantyo \textit{et al.} \cite{Sutantyo2013}, see Fig.~\ref{robot2}(a). 
Interaction between the robots was introduced by using a modification of
the Firefly Optimization, an algorithm popular in the field of particle swarm optimization \cite{swarm}. The ``attractiveness'' of 
each AUV  was defined 
by the time since the robot last found a target; it increased every time a target was located and then slowly decayed. 
The task was for each searcher 
to find all  the targets. 
The results of the experiments showed that the interaction  decreases the averaged search time substantially,
see Fig.~\ref{robot2}(b). 

Finally, an attempt to get insight into the machinery causing the emergence of L\'{e}vy walk-like  
patterns in the motion of different biological species was made recently by Fricke \textit{et al.} \cite{Fricke2013,Fricke2016}. Inspired by the results obtained for immune T-cells 
\cite{Harris2012}, researchers from the University of New Mexico and Santa Fe Institute 
used six small rovers,  equipped with ultrasound sensors, 
compasses, and cameras. This navigation set enabled  each robot to find patches of resources distributed over $2$-d area. 
Tunable adaptive algorithms  based on five different search strategies were tested. 
It turned out that the algorithm using correlated random walks, in which correlations between consequent step angles of a rover depend 
on the target last observed by the rover, produces L\'{e}vy-like motion patterns.

\section{Perspectives}\label{conclusion}

Here we want to share our feeling concerning the outgoing activity in the field of  L\'{e}vy robotics. 
But let us start from animals.

Any organism, even a bacteria, is a much more intellectual
being than a point-like particle driven by a finite-length
algorithm. From another point of view, ``a wandering albatross does  not
care about math'',  as it was perfectly noted by Travis \cite{travis}. It is 
naive to think that the albatross utilizes LWs when preying, 
by independently drawing a length of the next flight from a power-tailed probability distribution. 
Motion of a living organism is a product of a complex multi-layered activity of  
informational circuits  which are constantly processing external signals  
and generating internal signals  to control the organism's motion.  
An  anomalous dispersal pattern appears as the result of this  activity and not as a result of copycatting of some mathematical models. 

Therefore, the current activity looks to us a bit as a 'cargo cult' of LWs. 
Let us next to put the problem upside-down: If there are so many organisms, while
performing on so different time-range scales, produce motional patterns bearing one common and peculiar feature -- would it not be more reasonable to understand first
what is behind of it?  

Maybe (of course, it is a hope at the moment) there is some  simple mechanism, a kind of a sensory - locomotion loop which is responsible for the appearance of the  LW-like patterns.
The autonomous mobile robotics serves a perfect test-bed to validate (or refute)  any hypothesis. This path is  already taken -- but very recently -- in Refs.~\cite{Fricke2013,Fricke2016};
we do believe that it deserves more active research. There are some results from model simulations which suggest that LW-like patterns can be generated 
by a bacteria during a chemotactic activity because the bacteria is driven by a simple nonlinear circuit appearing due to chemotaxis-signaling pathway \cite{chem1, chem2}. 
Recent experimental observations demonstrated that swarming bacteria \textit{E. coli} also migrate ''by L\'{e}vy walk`` -- again,  because 
they chemically sense each other \cite{chem3}. To speculate further, we may think of the following conjecture:

\textit{Any white noise (Gaussian, shot noise, etc) when being 'filtered' through a system of a few nonlinear coupled differential equations and then used as an input to locomotion 
gears produces a motional pattern which could qualify (in some region of parameters) for  a 'Levy walk' (with tunable exponent $\gamma$) on a certain time scale.} 

This speculation (in case one was able to figure out a system of equations, a 'sensory loop')
can be checked in vitro, with a mobile robot, wheeled, legged  etc. Next, a tunable sensing can also be introduced  so that the robot is not only kicked by a spatially homogeneous noise but can
sense a target, though dimly. Then the activity of the robot can be continuously tuned from the task of locating a target to free-range exploration; see Fig.~ 4.

\begin{figure}[t]
\begin{tabular}{cc}
\includegraphics[width=0.48\textwidth]{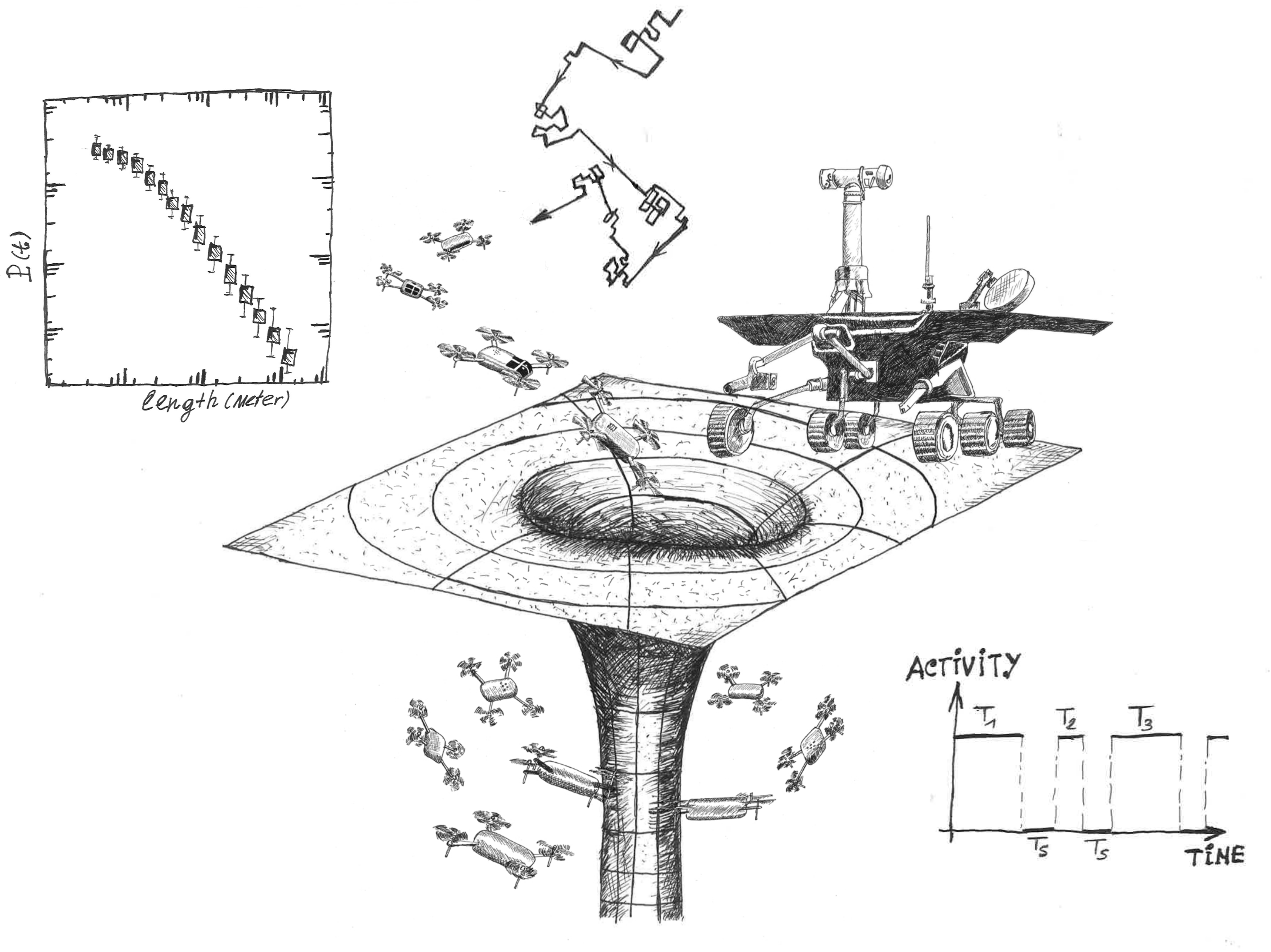}
\end{tabular}
\caption{An autonomous robot usually performs one of two activities: it either searches for a specified target(s) [a drone trying to localize
a gas emission source(s)] or performs free-range exploration of an unknown terrain [a Mars exploration rover]. The 'targeting' component of the robot activity
is visualized as an attractive potential field. Possible strategies for smart autonomous robots
can be thought as a continuous spectrum between the two extremes. }
\label{Fig:2}
\end{figure}

\section{Acknowledgments}\label{ac}

We are grateful to our  colleagues for discussions and suggestions:
C. Garcia-Saura, M. M\"{o}ckel,  M. Fricke, and  M. Timme. This work was supported by the the Russian 
Science Foundation grant No. 16-12-10496

\end{document}